\documentclass[journal]{IEEEtran}
\usepackage{amsfonts,amssymb,graphicx,subfigure,hyperref,booktabs,cite,setspace,multirow,mathrsfs}
\usepackage[ruled]{algorithm2e}

\begin{document}
\title{Graph Representation Learning via Contrasting Cluster Assignments}

\author{Chun-Yang Zhang,
	   Hong-Yu Yao,
        C. L. Philip Chen,~\IEEEmembership{Fellow,~IEEE}
        and~Yue-Na~Lin% <-this % stops a space
\thanks{Chun-Yang Zhang, Hong-Yu Yao (corresponding author), Yue-Na Lin are with College of Computer and Big Data, Fuzhou University, Fuzhou, China.}% <-this % stops a space
\thanks{C. L. Philip Chen is with School of Computer Science and Engineering, South China University of Technology, Guangzhou, Guangdong, 510006, China, and Navigation College, Dalian Maritime University, Dalian 116026, China.}% <-this % stops a space
\thanks{This research is sponsored in part by the National Natural Science Foundation of China under Grant No. 62076065, 61751202, 61751205, 61572540, U1813203, U1801262, and the Natural Science Foundation of Fujian Province under Grant No. 2020J01495.}}

% The paper headers
%\markboth{Journal of \LaTeX\ Class Files,~Vol.~11, No.~4, December~2012}%
%{Shell \MakeLowercase{\textit{et al.}}: Bare Demo of IEEEtran.cls for Journals}

% make the title area
\maketitle

\begin{abstract}
With the rise of contrastive learning, unsupervised graph representation learning has been booming recently, and shown strong competitiveness, even surpassing the supervised counterparts in some machine learning tasks. Most of existing contrastive models for graph representation learning either focus on maximizing mutual information between local and global embeddings, or primarily depend on contrasting embeddings at node level. However, they are still not exquisite enough to comprehensively explore the local and global views of network topology. Although the former considers local-global relationship, its coarse global information leads to grudging cooperation between local and global views. The latter pays attention to node-level feature alignment, so that the role of global view appears inconspicuous. To avoid falling into these two extreme cases, we propose a novel unsupervised graph representation model by contrasting cluster assignments, called as GRCCA, that can keep a balanced aggregation of local and global information. It is motivated to leverage clustering algorithms to grasp the more fine-grained global information (cluster-level), and get insight into the elusive association between nodes beyond graph topology. Moreover, we contrast embeddings at node level to preserve quality of local information, but enforce the cluster-level consistency instead of node-level consistency to explore global information elegantly. Specifically, we first generate two augmented graphs with distinct graph augmentation strategies, then employ clustering algorithms to obtain their cluster assignments and prototypes respectively. The proposed GRCCA further compels the identical nodes from different augmented graphs to recognize their cluster assignments mutually by minimizing a cross entropy loss. To demonstrate its effectiveness, we compare with the state-of-the-art models in three different downstream tasks, including node classification, link prediction and community detection. The experimental results show that GRCCA has strong competitiveness in most tasks.
\end{abstract}


\IEEEpeerreviewmaketitle

\section{Introduction}
\begin{figure*}[htbp]
	\centering
	\includegraphics[width=1\linewidth]{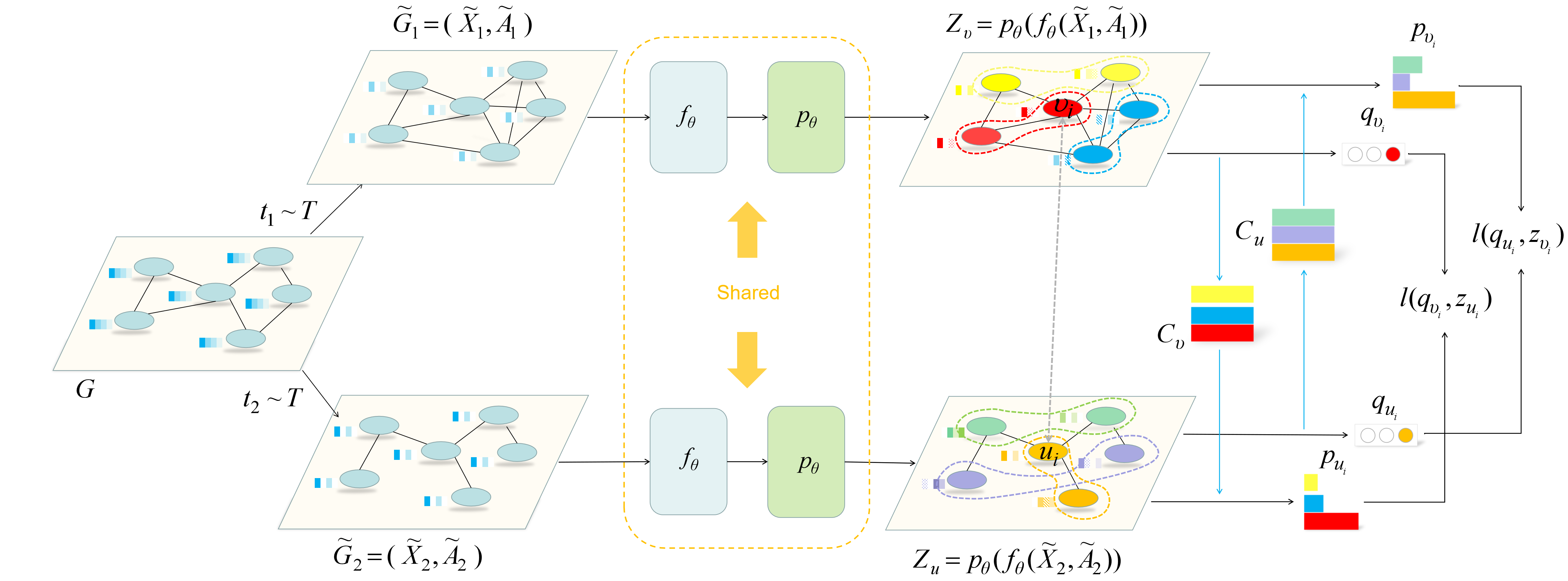}
	\caption{Illustration of the proposed unsupervised graph representation learning model by contrasting cluster assignments. First, two augmented graphs $\widetilde{\mathcal{G}}_1$ and $\widetilde{\mathcal{G}}_2$ are generated from the original graph $\mathcal{G}$ through the graph augmentations $t_1$ and $t_2$, which  are further fed into the encoder $f_\theta$ and the projector $g_\theta$ to generate two representation matrices $Z_v$ and $Z_u$. Then, two representation matrices are mapping into the prototypes $C_v$ and $C_u$ by k-means. Cluster assignments $Q_v$ and $Q_u$ indicate the cluster that each node belongs to after clustering, where $q_{v_i}$ and $q_{u_i}$ are the cluster assignments of the nodes $v_i$ and $u_i$ respectively. For any pair of the identical nodes $v_i$ and $u_i$, contrasting cluster assignments aims to make them recognize the other’s cluster category mutually, i.e the node $u_i$ are desired to be arranged to the cluster of node $v_i$ as much as possible and vice versa. The similarity between $u_i$ and $C_{v_i}$ are regarded as the predicted cluster distribution $p_{u_i}$, and then learning process can be achieved by minimizing the cross entropy loss between $p_{u_i}$ and $q_{v_i}$, which is same for the other side.}
	\label{GRCCA}
\end{figure*}

\IEEEPARstart {m}{any} real-world systems can be represented in the form of graphs or networks, such as social networks \cite{perozzi2014deepwalk, grover2016node2vec}, knowledge networks \cite{xie2016representation} and protein-protein intersection networks \cite{velickovic2019deep, peng2020graph, zhu2020deepc}. Hence, exploring knowledge from graph-structure data has great significance in many research fields, such as node classification, link prediction, community detection and so on \cite{hamilton2017representation}. However, the non-linearity and sparsity of graphs pose great challenges for effective knowledge discovery. Graph representation learning \cite{zhu2020deep, hamilton2017representation, zhang2019graph} is emerging as a powerful method to mine the graph-structure data, which learns low-dimensional and dense embeddings of nodes or graphs by refining topologies and attribute vectors of original graphs. Traditional graph representation models based on hand-craft features \cite{bhagat2011node, liben2007link} or matrix decomposition \cite{cao2015grarep, ou2016asymmetric} try to extract structural information from graphs, but they are too clumsy to adapt to realistic demand. Subsequently, random walk \cite{perozzi2014deepwalk, grover2016node2vec} provides a flexible and stochastic way to extract topological information. However, random walk based models still overly rely on structural proximity. Recently, with the prosperities of graph neural networks (GNNs) \cite{defferrard2016convolutional, kipf2016semi, velivckovic2017graph, wu2020comprehensive} and transformer-based models \cite{dwivedi2020generalization, ying2021transformers}, the problem has been solved partially. These models not only emphasize the structural information from network topology, but also make good use of node attributes. Although they have made great breakthroughs, most of them still need a certain number of labels as guidance. However, supervised learning has two disadvantages in nature. It requires a large volume of sample annotation, which may has human errors and is also time consuming. On the other hand, the representation generated by supervised models has poor generalization for other tasks. Therefore, unsupervised learning becomes another road leading to Rome.

The ingenious combinations of GNN with different unsupervised (or self-supervised) mechanisms expand the family of unsupervised graph representation learning. Auto-encoder as a typical unsupervised mechanism is naturally associated with unsupervised graph representation learning \cite{kipf2016variational, hasanzadeh2019semi}, while GraphSAGE \cite{hamilton2017inductive} inherits random walk mechanism as its unsupervised manner. Over the past few years, contrastive learning as a burgeoning unsupervised approach has received wide attention and achieved inviting results \cite{hjelm2018learning, he2020momentum, chen2020simple, grill2020bootstrap, caron2020unsupervised, ericsson2021well}. DGI \cite{velickovic2019deep} first employs contrastive learning to graph representation learning, and reproduces the success of contrastive learning in unsupervised visual representation learning \cite{hjelm2018learning}. Although DGI makes an important attempt, which contrasts embeddings in local-level and global-level representation by maximizing mutual information, its choice of the readout function can not get away from the harsh restriction. Consequently, the global-level representation obtained through this simple  averaging manner seems unconvincing, which may lose insight of differential characteristics between nodes. Nevertheless, MVGRL \cite{hassani2020contrastive} and GIC \cite{mavromatis2020graph} tenaciously seek to global-level information over graphs. The former enhances the contrastive mechanism of DGI with graph augmentation strategies and a new form of readout function, while the latter leverages cluster-level representations to refine the global-level information bridging the original gap between local and global relationships in DGI. To avoid falling into the trap of readout function, GMI \cite{peng2020graph} tries to maximize mutual information between the target node and its neighbors. What's more, there are some other successful models with various graph augmentation strategies and advanced experience in contrastive learning \cite{chen2020simple, grill2020bootstrap}, such as GRACE \cite{zhu2020deepc}, GCA \cite{zhu2021graph} and Merit \cite{jin2021multi}. They focus largely on feature alignments between the same nodes from multiple augmented perspectives, instead of utilizing local-global relationships.

However, the defects of the readout function do not mean that the global view is redundant. On the contrary, over-emphasis on the local node-level consistency may render a narrow understanding of graphs leading to another trap. To avoid this problem and make good use of the local-global relationship, inspired by SWAV \cite{caron2020unsupervised}, we propose a novel unsupervised Graph Representation learning model by Contrasting Cluster Assignments (GRCCA), which synthetically takes the global and local views into account. Clustering algorithms could naturally become an appropriate tool to bridge local and global views due to their inductive traits. For example, GIC makes an interesting attempt to refer to cluster-level information by explicitly emphasizing the consistency between the nodes and the corresponding cluster. Although it explores the fine-grained information of global view by clustering, it is not completely out of the shade of DGI and is insufficient to preserve distinctive node-level features. In contrast, the proposed GRCCA absorbs the respective advantages of the existing models. Concretely, we first employ two distinct graph augmentations to generate two augmented graph views $\widetilde{\mathcal{G}}_1$ and $\widetilde{\mathcal{G}}_2$ as shown in Fig. \ref{GRCCA}. The two augmented graphs focus on local and global views respectively, inducing the model to learn the invariance information from the contradiction. Then, we resort to a graph encoder $f_\theta$ and projector $g_\theta$ to obtain representation matrices $Z_v$ and $Z_u$ that are further used to obtain respective cluster assignments $Q_v,\,Q_u$ and prototypes $C_v,\,C_u$ by k-means. Finally, we enforce the identical nodes $v_i$ and $u_i$ from different perspectives to recognize their cluster assignments mutually by minimizing the cross entropy loss. In the contrastive clustering process, we implicitly consider the consistency between the nodes and the corresponding prototype, but design the contrastive objective at the node level. This further fuses clustering algorithms and contrastive learning to maximize the cooperation benefits between local and global views, rather than just enriches contrastive materials. In particular, the proposed GRCCA can be regarded as an eclectic between the models based on local-global contrastive objective and node-level feature alignment, where the former treats all nodes as a cluster, and the latter considers each node as a separate cluster.

Besides, we are no longer obsessed with the assumption that more proximate topologically more similar, while seeking to the inherent association between nodes through clustering. There may be some essential consistent patterns between the topologically remote nodes, while closely related nodes are quite different. It is just like people from different civilizations may have the same aesthetic pursuit, whereas people in the same civilization may breed different ideological sparks conversely. Our contributions are summarized as follows.
\begin{itemize}
\item A new graph representation learning model is proposed, which first employs the contrasting cluster assignments mechanism as an unsupervised manner, called GRCCA.
\item Through combining clustering algorithms and contrastive learning, GRCCA achieves a balance to explore global and local views, which not only facilitates the contrastive effect, but provides more comprehensively high-quality graph information.
\item GRCCA further excavates the cluster-level information in feature space to get insight into the abstract similarity between nodes. This similarity implies the essential pattern of graph and generalizes the assumption of proximate approximation and self-recognization.
\item Experimental results in three different tasks demonstrate the effectiveness of GRCCA. In node classification, GRCCA outperforms state-of-art unsupervised models and even some supervised models over six benchmark datasets. Moreover, it also remains sufficiently competitive in link prediction and community detection, while achieving the best results on some datasets. 
\end{itemize}

The rest of this paper is organized as follows. In Section \ref{relatedWorks}, some related works of unsupervised graph representation learning and clustering-based unsupervised learning are introduced. In Section \ref{methodDescription}, we describe our proposed GRCCA model in detail. The empirical results and comparative analysis are arranged in Section \ref{experimentsAndResults}. The conclusion and prospects for future work are drawn in Section \ref{conclusion}.

\section{Related work}
\label{relatedWorks}
The success of supervised learning is mainly due to large labeled corpora, but the pitfalls are always hidden behind a friendly facade \cite{jing2020self}, such as huge annotation costs, artificial annotation biases and so on. To overcome the label dependence problem, unsupervised methods are constantly innovating, from auto-encoder \cite{hinton2006reducing, kingma2013auto} to contrastive learning \cite{he2020momentum, chen2020simple, grill2020bootstrap, caron2020unsupervised} that gradually reduce the gap with supervised learning. Inspired by the success of different unsupervised (or self-supervised) mechanisms, various unsupervised graph representation learning models are proposed. By far, the most competitive models are mostly based on contrastive learning. According to the contrastive objectives, it can be roughly divided into three categories. The first category focuses on the local-global relationship \cite{velickovic2019deep, hassani2020contrastive, mavromatis2020graph}, the second category emphasizes the node-level mutual recognition between different perspectives \cite{zhu2020deepc, zhu2021graph, jin2021multi}, and the third one is committed to adjacency similarity \cite{peng2020graph}. In contrast, our model pays attention to the cluster-level consistency between the identical nodes by balancing the exploration between local and global views.

With the popularity of neural network, the long-dormant clustering algorithms are reawakened again \cite{caron2018deep, asano2019self, zhuang2019local, caron2019unsupervised, li2021contrastive}. As a natural unsupervised method, it can generate self-supervised signals in a way of assigning pseudo-labels to the training samples \cite{caron2018deep}, opening up a new way for unsupervised representation learning. To further excavate the potential of clustering algorithms, it is imperative to cooperate with contrastive learning \cite{zhuang2019local, caron2020unsupervised}. Clustering has become a popular way to construct positive and negative samples, providing more diverse materials for contrastive learning \cite{zhuang2019local,li2021contrastive}. SWAV \cite{caron2020unsupervised} highly integrates clustering algorithms with the idea of contrastive learning, which greatly promotes their cooperation benefits, and shows its unparalleled competitiveness in unsupervised visual representation learning. Recently, clustering algorithms have gradually begun to serve unsupervised graph representation learning, but they are still very scarce. Existing models focus largely on simple guidance from clustering pseudo-labels \cite{jin2020self, you2020does}, or try to improve the quality of contrastive materials \cite{mavromatis2020graph}. Different from them, we will take an ingenious way to fuse clustering algorithms and contrastive learning inspired by SWAV, providing a new choice for unsupervised graph representation learning.

\section{Proposed Graph representation learning via contrasting cluster assignments}
\label{methodDescription}
In this section, we elaborate the proposed GRCCA in two parts. The first part introduces the overall framework of GRCCA, including notations, graph augmentation strategies and model configuration. The second describes the learning algorithm in detail.

\subsection{Overall Framework}
In this paper, a graph is denoted as $\mathcal{G}=(\,\mathcal{V},\mathcal{E}\,)$, where $\mathcal{V}=\{{v_1},{v_2},\,...\,,{v_N}\}$ and $\mathcal{E}\in\mathcal{V}\times\mathcal{V}$ indicate the node set and the edge set respectively. Attribute matrix and adjacency matrix are represented by ${X}\in\mathbb{R}^{N\times F}$ and ${A}\in\{0,1\}^{N\times N}$, where ${x_i}\in\mathbb{R}^{F}$ denotes the attribute of $v_i$ and $A_{ij}=1$ iff $(v_i,v_j)\in\mathcal{E}$. Our target is to learn a graph encoder ${f_\theta}\,\colon\,\mathbb{R}^{N\times F}\times \mathbb{R}^{N\times N}\rightarrow\mathbb{R}^{N\times F^\prime}$ without any label signals, to create the low dimensional embeddings ${H}={f_\theta}({X},{A})\in\mathbb{R}^{N\times F^\prime}$, where $h_i$ denotes the embedding of node $v_i$ while $F^\prime\ll F$. The learned embeddings can be used in downstream tasks, such as node classification, link prediction and so on. 

Data augmentation strategies, such as rotation, scaling and cropping, are widely used in the visual domain, intending to increase sample diversity and generate multi-view information from limited samples. However, these strategies can not be directly applicable to graph-structure data due to isomorphism and non-Euclidean properties of graphs. Therefore, we adopt three currently popular graph augmentation strategies, including graph diffusion \cite{klicpera2019diffusion}, removing edges \cite{zhu2020deepc} and masking node features \cite{jin2021multi}, to enhance the diversity of graphs.

In the proposed GRCCA, Graph Diffusion (GD) and Removing Edges (RE) are used to produce two opposite perspectives that focus on global and local views of graph topology respectively. Graph diffusion removes the one-hop transmission of graph message, thus obtaining long-term dependencies of nodes information. It not only provides an augmented perspective with more global information, but also alleviates the problem of inborn noise in real graphs as an equivalent polynomial filter, which can be defined as
\begin{equation}\label{diffusion}
S=\sum_{k=0}^{\infty}\theta_k{T^{k}},
\end{equation}
where $T\in\mathbb{R}^{N\times N}$ is the generalized transition matrix to characterize the transfer form of the adjacency matrix, and $\theta_k$ is the weighting coefficient which determines the explored intensity of global information \cite{klicpera2019diffusion}. According to the empirical results of \cite{hassani2020contrastive}, we leverage Personalized PageRank (PPR) kernel as the instantiation of graph diffusion. Formally, given an adjacency matrix $A\in\mathbb{R}^{N\times N}$ and a diagonal degree matrix $D\in\mathbb{R}^{N\times N}$, PPR kernel can be given by
\begin{equation}\label{ppr}
S^\mathscr{ppr}=\alpha(I-(1-\alpha)D^{-1/2}AD^{-1/2}),
\end{equation}
where $I\in\mathbb{R}^{N\times N}$ is the identity matrix and $\alpha\in(0,1)$ is the teleport probability of random walk \cite{hassani2020contrastive}.
In contrast to graph diffusion, removing edges creates an opposite and stark perspective by reducing topological information. To tackle this, we follow the strategy in \cite{zhu2020deepc} to randomly remove the existing edges $e_{ij}$. Specifically, given an adjacency matrix $A$ and removal probability $P_{re}$, we randomly sample positive fractional number $b \sim U(0,1)$ with i.d.d hypothesis. The edge removal operation is performed if $b<P_{re}$ and $A_{ij}=1$.

For node-attribute-level augmentation, we perturb the graph attributes by masking nodes features (MNF) to obtain information from different perspectives. It is similar to cutOut strategy \cite{yun2019cutmix} in the visual augmentation, which aims to obtain the diverse perspectives of graph attributes through randomly masking. Specially, we employ the strategy in \cite{jin2021multi} to process attribute matrix. That is to randomly select masked attributes without repetition until the masked rate reaches the ratio $P_{mnf}$. It should be noted that each attribute dimension is selected by equal probability. Compared to the strategy in \cite{zhu2020deepc}, this strategy can facilitate the diversity of contrastive perspectives and enhance the robustness of learned embeddings as an anti-interference measure. Topology-level and node-attribute-level graph augmentation strategies not only provide multi-perspective knowledge on the graph, but also create contrast between them, like opposites between global and local views of graph topology, as well as significant differences between attribute matrices with random masks. Intuitively, the contrast forces the model to explore uniform invariance features from the two opposite augmented perspectives, thus increasing the effectiveness of contrastive learning.

After processing the original graph with data augmentation strategies, the two generated augmented graphs are passed to a shared graph encoder $f_\theta$ and nonlinear projector $g_\theta$ as shown in Fig. \ref{GRCCA}. For graph representation learning, the key is to preserve both structure and attributes information, so the choice of $f_\theta$ can theoretically be any encoders that takes both aspects into consideration. In this paper, we select the concise and efficient encoder GCN \cite{kipf2016semi} to get the nodes embeddings $H=f_\theta(\,X,A\,)$, accepting an adjacency matrix $A$ and an attribute matrix $X$ as input, which is defined as
\begin{equation}
H=\sigma(\hat{A}{X}{\Theta}),
\end{equation}
where $\hat{A}={\hat{D}^{-1/2}\hat{A}\hat{D}^{-1/2}}$ is the symmetrically normalized adjacency matrix, $\hat{D}$ is the degree matrix of the adjacency matrix with self-loop $\hat{A}=A+I_N$, $\Theta$ is the learnable parameters, $\sigma$ is the activation function. It is easy to construct deep structure by stacking GCNs. In order to enhance the expression ability of contrastive measure \cite{chen2020iterative, kaya2019deep}, we further use a nonlinear projector, i.e, MLP, to transfer node embedding to $Z=g_\theta(H)\in\mathbb{R}^{N\times F^\prime}$ in metric space. However, the success of this strategy is attributed to the combination of its adaptive generalization of metrics and its filtering effect on the information of excuse tasks. Analogy to deep neural networks, features in deeper layers are more appropriable to specified tasks than that in lower layers. This may lead to some negative impacts on other tasks. To reduce the amount of model parameters, the encoder and the projector share the same parameters for different augmented graphs. 
%The additional projector is equivalent to increasing the depth of network, aggregating the task-specific information of contrastive learning into the projector to ensure generalization capability of the learned node embeddings.

\subsection{Learning Algorithm}\label{LearningAlgorithm}
Different from previous graph contrastive learning models and clustering-based graph representation learning models, GRCCA combines contrastive learning and clustering algorithms together by maximizing cluster-level consistency between the identical nodes from two augmented perspectives. Contrasting cluster assignments not only increases the cooperative benefits between contrastive learning and clustering algorithms, and furthermore provides a desirable way to reconcile local and global views. Once we obtain two representation matrices $Z_v$ and $Z_u$ by passing two augmented graphs through the encoder $f_\theta$ and projector $g_\theta$ as shown in Fig. \ref{GRCCA}, k-means is further applied to generate two prototypes $C_v\in\mathbb{R}^{K\times F^\prime}$ and $C_u\in\mathbb{R}^{K\times F^\prime}$ from the two representation matrices, where $K$ is the number of clusters and $F^\prime$ indicates the dimension of prototypes. According to the cluster to which each node representation belongs, we can get two cluster assignments $Q_v$ and $Q_u$, where each entry $q_{v_i}=\mathbb{I}_{z_{v_i}\in C_{v}^K}$ or $q_{u_i}=\mathbb{I}_{z_{u_i}\in C_{u}^K}$ indicates one-hot cluster category of nodes. Rather than assigning pseudo-labels to themselves in common clustering-based models, the proposed GRCCA compels the identical nodes of another perspective to recognize the cluster assignments of current perspective by minimizing the cross entropy loss. For any pair of the identical nodes $v_i$ and $u_i$, their contrastive loss can be given by
\begin{equation}
\ell(q_{v_i},z_{u_i})=-q_{v_i}\log{p_{u_i}},
\end{equation}
\begin{equation}
	p_{u_i}=softmax(\frac{z_{u_i}{C_v}^\mathrm{ T }}{\tau}),
\end{equation}
where $p_{u_i}$ is the prediction cluster label of the identical node from first augmented graph, and $\tau$ is a temperature parameter. For another augmented graph, it is easy to obtain a symmetrical loss $\ell(q_{u_i},z_{v_i})$ that has the same form. Formally, we can define the contrastive over all the nodes by integrating above two losses as
\begin{equation}
\mathcal{L}_{c}=\frac{1}{N}\sum_{i=0}^{N}\,[\;\ell(q_{v_i},z_{u_i})+\ell(q_{u_i},z_{v_i})\;].
\end{equation}
Contrasting cluster assignments can be regarded as a special way of contrastive learning, which compares cluster assignments between multiple graph perspectives rather than their features. It implicitly drives the node embeddings to approach their corresponding prototypes, and set apart from other prototypes. Intuitively, it is equivalent to maximizing the mutual information between the node embeddings and the corresponding prototypes.

Inspired by the multi-head attention mechanism \cite{vaswani2017attention}, GRCCA employs a multi-clustering strategy to increase the diversity of cluster-level information. Specifically, we perform multiple clustering for each perspective synchronously, yielding the multiple pairwise contrastive materials $\{(C_{v}^{1},C_{u}^{1},Q_v^{1},Q_u^{1}),\,...\,,(C_{v}^{h},C_{u}^{h},Q_v^{h},Q_u^{h})\}$, and utilize the contrasting cluster assignments to ensure their cluster-level consistency. Therefore the totally loss $\mathcal{L}_{mc}$ can be given by
\begin{equation}
\mathcal{L}_{mc}=\frac{1}{h}\sum_{i=0}^{h}\,\mathcal{L}_{c}^{i},
\label{lmsc}
\end{equation}
where $h$ denotes the number of contrastive material sets.

The learning algorithm is summarized in Algorithm \ref{algrm}. Firstly, we apply two graph augmentation function $t_1\sim T$ and $t_2\sim T$ to generate two augmented graphs $\widetilde{\mathcal{G}}_1$ and $\widetilde{\mathcal{G}}_2$, where $t_1$ consists of $GD$ and $MNF$, and $t_2$ includes $RE$ and $MNF$. Secondly, we employ an graph encoder $f_\theta$ and nonlinear projector $g_\theta$ to generate the node representations of two views respectively. Thirdly, k-means with multi-clustering strategy $K_{m}$ is used to generate cluster assignments $Q_v$, $Q_u$ and prototypes $C_v$, $C_u$. Fourthly, we minimize the contrastive loss in Eq.  [\ref{lmsc}] to enforce the cluster-level consistency between the identical nodes from different perspectives. Otherwise, we try two different cluster assignments scheme: asynchronous and synchronous versions. Asynchronous version uses the representation matrices from the previous epoch to generate cluster assignments, while synchronous version refers to the current representation matrices. Notably, the asynchronous version needs to initialize a memory bank $B$ and update it with each round of representation. Finally, the node embeddings learned from the graph encoder $f_\theta$ are used for downstream tasks.  
\begin{algorithm}[h]\label{training}
  \SetAlgoLined
  \KwIn{graph $\mathcal{G}=\{\,X,A\,\}$, parameters $\Theta$, learning rate $\beta$}
  \KwResult{node embeddings $H$}
  \setstretch{1.25}
  $\widetilde{\mathcal{G}}_1 = t_{1}(\mathcal{G});$\\
  $\widetilde{\mathcal{G}}_2 = t_{2}(\mathcal{G});$\\
  $H_v = f_\theta(\widetilde{\mathcal{G}}_1);$\\
  $H_u = f_\theta(\widetilde{\mathcal{G}}_2);$ \\
  $B \leftarrow H_v,\,H_u;$ \\ \tcp{\footnotesize{initialize memory bank}}
  \For{epoch$\,=\,$$1\,:\,T$}{
	\tcp{\footnotesize{asynchronous cluster assignments}}
	$C_{v},C_{u},Q_v,Q_u \leftarrow K_{m}(B);$  \\
	$\widetilde{\mathcal{G}_1} = t_{1}(\mathcal{G});$\\
  	$\widetilde{\mathcal{G}_2} = t_{2}(\mathcal{G});$\\
  	$H_v = f_\theta(\widetilde{\mathcal{G}}_1);$\\
  	$H_u = f_\theta(\widetilde{\mathcal{G}}_2);$ \\
	$Z_v = g_\theta(H_v);$\\
	$Z_u = g_\theta(H_u);$\\
	\tcp{\footnotesize{compute contrastive loss in Eq.[\ref{lmsc}]}} 
	$\mathcal{L}_{mc} \leftarrow C_{v},C_{u},Q_v,Q_u,Z_v,Z_u;$\\
	$\Theta_{t} = \Theta_{t-1} - \beta\bigtriangledown\mathcal{L}_{mc};$ \tcp{\footnotesize{update parameters}}
	$B \leftarrow Z_v,\,Z_u;$  \tcp{\footnotesize{update memory bank}}
  }
  $ H = f_\theta(\mathcal{G}) + f_\theta(GD(\mathcal{G}));$\\
  \textbf{return}{$\;\;H;$}
  \caption{Learning algorithm for GRCCA}
\label{algrm}
\end{algorithm}

\section{Experiments}
\label{experimentsAndResults}
To evaluate the effectiveness of the proposed GRCCA, we present a comparative analysis with other baseline models on three different graph tasks in this section.  The GRCCA is implemented by PyTorch over six datasets, and all experiments are conducted on workstations configured as follows
\begin{itemize}
\item CPU: Intel(R) Xeon(R) Silver 4110 CPU @ 2.10GHz *2, Memory: 32G, GPU: GeForce GTX 1080 *2.
\item CPU: Intel 10900X @3.70 GHz~4.50 GHz(10C20T), Memory: 32G, GPU RTX3060*2.
\end{itemize}

\subsection{Datasets}
To comprehensively evaluate the performance of the proposed GRCCA, we use six popular datasets from different domains, including Cora, Citeseer, Pubmed, Amazon-Photo, Amazon-Computers and Coauthor-CS. The relevant statistics of datasets are shown in Table \ref{dataset}.
\begin{table}[htbp]
	\renewcommand\arraystretch{1.1}
	\caption{The statistics of datasets} 
	\begin{tabular}{ccccc}    
		\toprule    
		Dataset & Nodes & Edges & Features & Class\\    
		\midrule   
		Cora & 2,708 & 5,429 &  1,433 & 7\\   
		Citeseer & 3,327 & 4,732 & 3,703 & 6\\   
		Pubmed &  19,717 & 44,338 &  500 & 3 \\    
		Amazon-Photo & 7,650 & 119,081 & 745 & 8 \\    
		Coauthor-CS & 18,333 & 81,894 & 6,805 & 15\\    
		Amazon-Computers & 13,752 & 245,861 & 767 & 10 \\    
		\bottomrule   
	\end{tabular}  
	\label{dataset}
\end{table}

\textbf{Cora}, \textbf{Citeseer} and \textbf{Pubmed} are citation networks that record citing or cited relationship between papers (or publications). The node attribute vector indicates whether the paper (or publication) contains the specified keywords and class labels indicate different categories.

\textbf{Amazon-Photo} and \textbf{Amazon-Computers} are two co-purchase graphs collected by crawling Amazon website, in which nodes represent goods, edges indicate that two goods are frequently bought together, node attributes are bag-of-words encoded product reviews and class labels represent their product categories. \cite{shchur2018pitfalls}.

\textbf{Coauthor-CS} is a co-authorship graph based on the Microsoft Academic Graph from the KDD Cup 2016 challenge, where nodes represent authors, two nodes connected by an edge mean they are co-authors in a paper. Node attributes are derived from the paper keywords of each author's papers, and class labels are the most active research domains studied by each author \cite{shchur2018pitfalls}.

%%%%
\subsection{Node Classification}\label{nodeClassification}

\begin{table*}[htbp]
	\centering
	\renewcommand\arraystretch{1.2}
	\caption{The hyperparameter settings}   
	\begin{tabular}{cccccccc}    
		\toprule    
		Dataset & $P_{re}$ & $P_{nmf}^{1}$ & $P_{nmf}^{2}$ & $\tau$ & $N_{pt}$ & $\beta$ & $\sigma$\\    
		\midrule   
		Cora & 0.2 & 0.3 & 0.4 & 0.05 & 14 & 0.0005 & ReLU\\   
		Citeseer & 0.2 & 0.3 & 0.4 & 0.1 & 21 & 0.0005 & PReLU\\
		Pubmed & 0.4 & 0.2 & 0.2 & 0.15 & 9 & 0.001 & ReLU\\
		Amazon-Photo & 0.4 & 0.4 & 0.3 & 0.15 & 17 & 0.001 & ReLU\\
		Coauthor-CS & 0.4 & 0.2 & 0.2 & 0.1 & 24 & 0.001 & ReLU\\
		Amazon-Computers & 0.4 & 0.2 & 0.2 & 0.1 & 21 & 0.001 & PReLU\\
		\bottomrule   
	\end{tabular}  
	\label{hyperparameter}
\end{table*}
\begin{table*}[htbp]
	\renewcommand\arraystretch{1.3}
	\caption{Experimental performance of node classification}   
	\begin{tabular}{cccccccc}    
		\toprule    
		Model & Training Data & Cora & Citeseer & Pubmed & Amazon-Photo & Coauthor-CS & Amazon-Computers \\    
		\midrule   
		Chebyshev & \textbf{X, A, Y} & 79.3 &  68.0 & 74.0 & 86.5 & 90.2 & 74.4\\   
		GCN & \textbf{X, A, Y} & 81.5 & 70.3 & 79.0 & 85.8 &	92.4	& 75.3\\   
		GAT & \textbf{X, A, Y} & 83.7$\pm$0.7 & 72.5$\pm$0.7 & 79.0$\pm$0.3 & 84.8$\pm$0.5 & 90.9$\pm$0.1 & 76.0$\pm$0.1\\    
		\midrule 
		
		DGI & \textbf{X, A} & 81.8$\pm$0.2 & 71.5$\pm$0.2 & 77.3$\pm$0.6 & 80.5$\pm$0.2	& 91.0$\pm$0.0 & 75.3$\pm$0.1\\    
		GMI & \textbf{X, A} & 82.7$\pm$0.2 & 73.0$\pm$0.3 & 80.1$\pm$0.2 & 83.9$\pm$0.0 & 91.7$\pm$0.0 & 78.2$\pm$0.0\\    
		GIC & \textbf{X, A} & 82.7$\pm$0.2 & 70.0$\pm$0.4 & 79.1$\pm$0.2 & 87.1$\pm$0.1 & 91.7$\pm$0.0 & 78.7$\pm$0.1\\    
		GCA & \textbf{X, A} & 83.1$\pm$0.3 & 72.3$\pm$0.5 & 80.2$\pm$0.3 & 89.4$\pm$0.1 & 91.8$\pm$0.0 & 81.5$\pm$0.2\\ 
		MVGRL & \textbf{X, A} & 84.7$\pm$0.7	& 73.7$\pm$0.1 & 80.8$\pm$0.1 & 85.8$\pm$0.5 & 89.6$\pm$0.1 & 79.3$\pm$0.3\\
		MERIT & \textbf{X, A} & 83.5$\pm$0.2 & 74.0$\pm$0.3 & 80.8$\pm$0.2 & 87.5$\pm$0.3 & 92.6$\pm$0.1 & 81.3$\pm$0.3\\
		\textbf{GRCCA} & \textbf{X, A} & \textbf{85.1$\pm$0.1} & \textbf{74.5$\pm$0.1} & \textbf{81.7$\pm$0.1} & \textbf{91.7$\pm$0.2} & \textbf{93.0$\pm$0.2} & \textbf{85.0$\pm$0.1}\\
		\bottomrule   
	\end{tabular}  
	\label{classification}
\end{table*}
Node classification is a classical task used to test the abilities of graph representation learning models. To illustrate the effectiveness of our model, we select six state-of-the-art unsupervised models including DGI \cite{velickovic2019deep}, GMI \cite{peng2020graph}, MVGRL \cite{hassani2020contrastive}, GIC \cite{mavromatis2020graph}, GCA \cite{zhu2021graph} and MERIT \cite{jin2021multi}, and three classical GNN models: ChebyshevGCN, GCN, and GAT \cite{kipf2016semi, velivckovic2017graph} as baselines. In order to ensure the reliability of the results, we implement their published codes, and fine tune their models over same training datasets splited with a consistent scheme as \cite{jin2021multi}. Concretely, for three citation networks we randomly sample $20$ nodes per class to form a training set and $1000$ nodes as testing set. While for other three datasets, we randomly select $30$ nodes per class to construct the training and validation sets, while the remaining nodes are used for testing. For each experiment, all the models are firstly trained in unsupervised manner. Then the learned node embeddings are further employed to accomplish the node classification task by a naive linear classifier. Besides, we run $10$ times experiments independently to obtain mean accuracy with standard deviation referring to \cite{jin2021multi}.

To reduce the amount of model parameters, we use one-layer GCN and two-layer MLP with batch normalization as the backbone of GRCCA, which is applied independently to two augmented graphs with shared parameters for all tasks. The training process adopts Adam optimizer to update gradients. For all datasets, the dimension of the encoder and projector are fixed to $256$, the teleport probability of graph diffusion $\alpha$ is $0.05$ and the number of multi-clustering $h$ is set to $2$. The number of iterations is $10$ on all data sets except for $20$ on Amazon-Computers. Other hyperparameter are shown in Table \ref{hyperparameter}, including the rate of removing edges $P_{re}$, the rates of masking nodes features $P_{nmf}^{1}$ and $P_{nmf}^{2}$, temperature parameter $\tau$, the number of prototypes $N_{pt}$, learning rate $\beta$ and activation function $\sigma$. To select a relatively optimal model, we adopt the grid searching within a certain range, that is, to select $P_{re}$, $P_{nmf}^{1}$ and $P_{nmf}^{2}$ in $\{0.1, 0.2, 0.3, 0.4\}$, $\tau$ in $\{0.05, 0.1, 0.15, 0.2\}$ and $N_{pt}$ in the integer range between one and four times the number of original label classes. 

\begin{table*}[htbp]
\centering
\renewcommand\arraystretch{1.3}
\caption{The ablation study}   
\begin{tabular}{ccccccc}    
\toprule    
Model & Cora & Citeseer & Pubmed & Amazon-Photo & Coauthor-CS & Amazon-Computers \\    
\midrule 
\textbf{GRCCA} & \textbf{85.1$\pm$0.1} & \textbf{74.5$\pm$0.1} & \textbf{81.7$\pm$0.1} & \textbf{91.7$\pm$0.2} & \textbf{93.0$\pm$0.2} & \textbf{85.0$\pm$0.1}\\
GRCCA$/$MC & 84.5$\pm$0.1 & 73.9$\pm$0.2	& 81.6$\pm$0.1	& 91.5$\pm$0.1	& \textbf{93.0$\pm$0.0} & 84.6$\pm$0.1\\
GRCCA$/$AM & 83.7$\pm$0.2 & 74.0$\pm$0.1 & 76.9$\pm$0.1 & 91.6$\pm$0.1	& 92.8$\pm$0.1	& \textbf{85.0$\pm$0.1}\\
GRCCA$/$GD & 82.7$\pm$0.5 & 73.8$\pm$0.1	 & 80.7$\pm$0.3 & 91.1$\pm$0.2	& 92.4$\pm$0.2 & 83.7$\pm$0.3\\
\bottomrule   
\end{tabular} 
\label{ablation} 
\end{table*}
\begin{table*}[htbp]
	\centering
	\renewcommand\arraystretch{1.3}
	\caption{The number of learnable parameters}   
	\begin{tabular}{ccccccc}    
		\toprule    
		Model & Cora & Citeseer & Pubmed & Amazon-Photo & Coauthor-CS & Amazon-Computers \\    
		
		\midrule 
		
		DGI & 996,354 & 2,158,594 & \textbf{193,794} & 644,098 & 3,746,818 & 655,362\\    
		GMI & 1,992,709 & 4,054,532 & 774,660 & 1,025,540 & 7,230,980 &	1,048,068\\    
		GIC & 996,354 & 2,158,594 & \textbf{193,794} & 644,098 & 3,746,818 & 655,362\\    
		GCA & 997,120 & 2,159,361 & 519,424 & 546,368 & 3,747,584 & \textbf{256,896}\\ 
		MVGRL & 1,730,563 & 4,055,043 & 775,171 & 1,026,051 & 7,231,491 & 1,048,579\\
		MERIT & 14,089,733 & 16,414,213 & 13,134,341 & 13,385,221 & 19,590,661 & 13,407,749\\
		\textbf{GRCCA} & \textbf{499,201} & \textbf{1,080,322} & 260,353 & \textbf{323,073} & \textbf{1,874,433} & 344,834\\
		\bottomrule   
	\end{tabular}  
	\label{parameter}
\end{table*}
Table \ref{classification} shows the classification accuracy of respective models over six datasets, where the proposed GRCCA outperforms both unsupervised models and the classical models with label supervision on all datasets, particularly by $2.3\%$ and $3.5\%$ than the sub-optimal models on the two co-purchasing networks, i.e Amazon-Photo and Amazon-Computers. This may be simply because GRCCA comprehensively explores local and global views and grasps the inherent association between nodes by clustering. Furthermore, there are other observations from the experimental results that justify our motivation. From the comparisons with DGI and GIC, it is safe to conclude that the exploration of cluster-level information in graph makes sense. Although GIC as an expedient whose performance is not inconspicuous compared to other models, by appending the exploration of cluster-level information into DGI, it does make use of global information more efficiently and show more competitive results than DGI. Besides, it is also obvious that two types of models represented by MERIT and MVGRL show strong competitiveness in node classification, where node-level consistency is slightly dominant. While the benefits of global view seem inconspicuous on the surface, this may have more to do with how global information is explored. From another perspective, MVGRL just simply relies on such crude global view, but still induces such sufficient competitiveness. This motivates us to recognize the indispensability of the global view. However, we cannot say turkey to one and buzzard to another, and put aside the great success of the models that primarily focus on node-level information in local view. Therefore, we propose GRCCA to conciliate two categories for equilibrium points in the two schemes. The experimental results also support the validity of the motivation. From Table \ref{classification}, it is interesting to find that the gap between unsupervised and supervised models is gradually narrowed, and unsupervised models can even prevail as contrastive learning and graph augmentation strategies to be mature. This is further improved by contrasting cluster assignments in GRCCA.

\begin{table*}[htbp]
\centering
\renewcommand\arraystretch{1.3}
\caption{Experimental performance of link prediction}
\begin{tabular}{ccccccc}
\toprule  

 \multirow{2}{*} {Model} &

 \multicolumn{2}{c}{Cora} &

 \multicolumn{2}{c}{Amazon-Photo} &

 \multicolumn{2}{c}{Coauthor-CS} \\

 \cline{2-7} 
   & AUC & AP & AUC & AP & AUC & AP \\
\midrule 

DGI & 95.41$\pm$0.08 & 95.45$\pm$0.04 & 92.24$\pm$0.09 & 92.14$\pm$0.06 & 94.87$\pm$0.07 & 94.34$\pm$0.09 \\
GMI & 95.58$\pm$0.25 & 95.71$\pm$0.44 & 93.88$\pm$1.00 & 92.67$\pm$1.43 & 96.37$\pm$0.46 & 96.04$\pm$0.55 \\
GIC & 95.75$\pm$0.33 & 95.66$\pm$0.28 & 92.70$\pm$0.95 & 92.34$\pm$0.47 & 95.03$\pm$0.12 & 94.94$\pm$0.19 \\
GCA & 95.75$\pm$0.13 & 95.47$\pm$0.12 & 93.25$\pm$0.71 & 92.74$\pm$0.24 & 96.31$\pm$0.15 & 96.28$\pm$0.11 \\
MVGRL & 90.52$\pm$0.28 & 90.45$\pm$0.78 & 92.89$\pm$0.17 & 92.45$\pm$0.42 & 95.17$\pm$0.11 & 95.08$\pm$0.13 \\
MERIT & \textbf{96.77$\pm$0.12} & 96.69$\pm$0.09 & \textbf{96.27$\pm$0.23} & \textbf{95.84$\pm$0.18} & 97.23$\pm$0.05 & 97.27$\pm$0.08 \\
\textbf{GRCCA} & 96.76$\pm$0.10 & \textbf{96.78$\pm$0.15} & 96.23$\pm$0.11 & 95.43$\pm$0.14 & \textbf{97.61$\pm$0.09} & \textbf{97.68$\pm$0.08} \\

\bottomrule  

 \end{tabular}
\label{link} 
\end{table*}

To better investigate the impacts of respective schemes in our model, we carry out ablation studies as shown in Table \ref{ablation}, where GRCCA$/$MC denotes GRCCA without multi-clustering strategy, GRCCA$/$AM represents GRCCA without asynchronous mechanism, i.e., synchronous version mentioned in Section \ref{LearningAlgorithm} which applies current representation to achieve the contrastive clustering, and GRCCA$/$GD is GRCCA without graph diffusion strategy. From Table \ref{ablation}, it is easy to see that multi-clustering strategy brings a slight boost on two smaller citation networks and negligible effects on the larger datasets. This may be because k-means can enrich the diversity of contrast materials for small-scale datasets. However, for large-scale graphs, there are sufficient cluster samples so that the clustering results are relatively stationary, thus weakening the uncertainty of k-means and making the multi-clustering strategy redundant.

The original incentive of using the asynchronous mechanism is to make the training process more continuous, so that the learned embeddings will be more robust. In the three citation networks, the mechanism shows  effects as envisioned, particularly on Pubmed. It is probably attributed to the node attributes of the citation networks, which are  indicator vectors of specific words. It is difficult for k-means to get a relatively fixed clustering result, so that the model has to adapt to the current contrastive clustering task from scratch in each iteration. The other three datasets use the word-bag vectors as their node attributes that have a relatively large dimension. They are more dense, thus containing more information. A sufficient number of nodes also provides rich clustering samples for them. So k-means can obtain a relatively stable result that ensures the continuity of GRCCA. Obviously, graph diffusion enhances its performance and gives it more global information. In addition, it is both contradictory and complementary to removing edges, facilitating the model to reconcile local and global information, and also getting insight into the underlying invariance traits.

It is also  interesting  to investigate the complexity of  proposed GRCCA, so that we also summarize the number of learnable parameters of respective models to show their conciseness and efficiency. As illustrated in Table \ref{parameter}, it is obvious that the number of GRCCA's parameters are minimal over almost every datasets. It reduces nearly half the parameters than the second lowest model, except for Pubmed and Amazon-Computer. With fewer parameters, GRCCA does not struggle to cater to computational power, and has lower risk of overfitting to some extent. Totally, it indirectly reflects its efficiency, i.e., GRCCA obtain more significant effects with less parameters.

\begin{table*}[htbp]
\centering
\renewcommand\arraystretch{1.3}
\caption{Experimental performance of community detection}
\setlength{\tabcolsep}{4.5mm}{
\begin{tabular}{cccccccccc}
\toprule  

 \multirow{2}{*} {Model} &

 \multicolumn{3}{c}{Cora} &

 \multicolumn{3}{c}{Citeseer} &

 \multicolumn{3}{c}{Pubmed} \\

 \cline{2-10} 
   & ACC & NMI & ARI & ACC & NMI & ARI & ACC & NMI & ARI \\
\midrule 

DGI & 71.5 & 55.9 & 50.6 & 68.3 & 43.9 & 44.2 & 66.8 & 30.4 & 28.4 \\
GMI & 70.7 & 55.5 & 47.2 & 66.6 & 41.7 & 42.3 & 64.0 & 25.8 & 23.6 \\
GIC & 73.7 & 56.1 & 53.2 & 69.9 & \textbf{45.2} & 46.3 & 67.3 & 31.7 & 29.1 \\
GCA & 71.1 & 55.7 & 47.8 & 63.7 & 37.4 & 37.5 & 65.2 & 28.9 & 27.8 \\
MVGRL & \textbf{76.7} & \textbf{60.9} & \textbf{56.8} & 69.4 & 44.0 & 44.2 & 66.7 & 31.5 & 28.8 \\
MERIT & 75.1 & 59.6 & 56.6 & 68.3 & 43.8 & 43.3 & 67.7 & 33.8 & 30.1 \\
\textbf{GRCCA} & 73.4 & 59.6 & 53.4 & \textbf{70.2} & 45.1 & \textbf{46.4} & \textbf{69.5} & \textbf{34.0} & \textbf{32.5} \\

\bottomrule  

 \end{tabular}}
\label{detection} 
\end{table*}

\subsection{Link Prediction}
For link prediction task, models are required to predict the presence of edges between nodes with incomplete topological information. In the following experiments, the six unsupervised models mentioned in the node classification task are selected for comparison over Cora, Amazon-Photo and Coauthor-CS networks. According to \cite{mavromatis2020graph}, the training set is constructed by removing parts of edges, while validation and test sets are formed from previously removed edges and the same number of randomly sampled pairs of unconnected nodes. The validation set contains $5\%$ of links from the original graph, and test sets contain $10\%$ of connections. For each model, we compare on the same divided datasets. The mean and variance of their area under the ROC curve (AUC) and average precision (AP) scores are used as evaluative criteria after five independent experiments over each dataset. Because some competitors have not attempted the link prediction task, it is irresponsible to directly evaluate their performance without any adjustments. For fairness, we try our best to adjust their hyperparameters and report their best results as comparative reference. Meanwhile, we slightly adjusted the hyperparameters of GRCCA to better adapt to the link prediction task. 

As shown in Table \ref{link}, GRCCA also demonstrates some competitiveness in link prediction task. Although it is slightly inferior to MERIT on Amazon-Photo, it has a great advantage over other models. From Table \ref{link}, it is easy to see that models focusing on local view perform better overall. Intuitively, this result is expected because link prediction task seems more concerned with local neighborhood. Despite the performance of models considering global-level information is not excellent in experiments, this can not deny the role of global-level information, which is not an unfounded judgment. For example, MERIT clearly outperforms other same type models in experiments, and interestingly it explores global-level information through graph diffusion in the data augmentation, while the other models do not make such attempts. Moreover, although GRCCA retains node-level contrastive objective to grasp more exquisite local-level information, it also takes global-level information into account, which even makes it outperform MERIT on Coauthor-CS. Therefore, it is reasonable to believe that global information is not redundant for the link prediction task. In other words, the factors that really affect the performance of  models may be the way they explore global-level information like readout function rather than global-level information itself. 

\subsection{Community Detection}
For community detection task, three citation network are selected for experiments that aim to detect the research areas (communities) to which each article belongs, where the nodes are in the same community if they have the same label. We still choose the six unsupervised competitors to compare with GRCCA, and employ k-means to cluster their node embedings into different communities for each model. Similar to link prediction task, we try our best to adjust competitors’ hyperparameters and make them adapt to community detection in order to ensure the experiments as fair as possible. The hyperparameters of GRCCA are also fine-tuned to fit community detection task. Classification accuracy (ACC), normalized mutual information (NMI) and average rand index (ARI) are chosen as evaluative criteria \cite{mavromatis2020graph} to validate the ability of learned embeddings. 

The experimental results of community detection are reported in Table \ref{detection}, in which GRCCA achieves better results on Citeseer and Pubmed. Specifically, GRCCA outperforms other models on Pubmed in a big margin, but performs equally on Cora. Similar results are presented for the GIC, it may be because both of them employ clustering algorithms. It is easy to find that their advantages are relatively obvious on networks like Citeseer that have very sparse topology and isolated points, as the clustering algorithm provides more detailed global information and excavates essential associations between nodes. However, clustering may also bring some negative effects. For example, two nodes do not belong to the same community, but there may be some similarity between them, so that causes the model to learn some information that is not relevant to the community detection task. Consequently, GRCCA does not perform as well on Cora than MERIT and MVGRL. Besides, models focusing on global view generally perform well, while models like GMI and GCA that pay more attention to local-level view perform relatively inferior. This phenomenon is reasonable because each model has its own preference, and community detection is intuitively more dependent on global information. As for the reason why MERIT still remains competitive, it may be related to the superiority of its own structure \cite{grill2020bootstrap} in addition to its consideration of global information. Totally speaking, the proposed GRCCA could be an alternative and competitive unsupervised graph representation model.

\section{Conclusion}
\label{conclusion}
To better integrate global and local information, we propose a novel unsupervised graph representation learning model by contrasting cluster assignments, called as GRCCA. In addition to balancing the global and local views, it explores the cluster-level information contained in graphs so as to get insight into potential associations between nodes rather than just topological proximity. Furthermore, the proposed GRCCA shows a new avenue for unsupervised graph representation learning by combining clustering algorithms with contrastive learning to promote the cooperation between local and global views. Meanwhile, it shows that cluster-level information has great mining potential, which can excavate more exquisite global information from graphs. Experimental results also confirmed the rationality of our motivation and the effectiveness of GRCCA.

Although the rapid development of graph representation learning is exciting, there are still many problems to be solved, such as the scale of graphs, the evolution of real-world graphs, graph-level representation learning and so on. Generally, the scale of real-world networks is very huge, which challenges both computational power and the processing power of models. What’s more, they are also constantly evolving, which involve not only structural information, but also temporal evolving information \cite{kazemi2020representation, skardinga2021foundations}. Existing computing power cannot solve these problems directly, so we have to conceive a more effective blueprint to satisfy practical application. How to incrementally learn graph-structure data and mine the dynamics in real networks will be the direction of our future research. Besides, most existing methods focus on node-level representation learning, while graph-level representation learning is also crucial \cite{zeng2020contrastive, sun2021sugar}. To the best of our knowledge, the attempts at combination between clustering algorithms and graph-level representation learning remain blank. Most importantly, the clustering prototypes can be regarded as a fine-grained graph-level representation, so extending our idea to the graph-level representation learning is a promising direction.

\ifCLASSOPTIONcaptionsoff
  \newpage
\fi

\bibliographystyle{unsrt}
\bibliography{cite}

\begin{IEEEbiography}[{\includegraphics[width=1in,height=1.25in,clip,keepaspectratio]{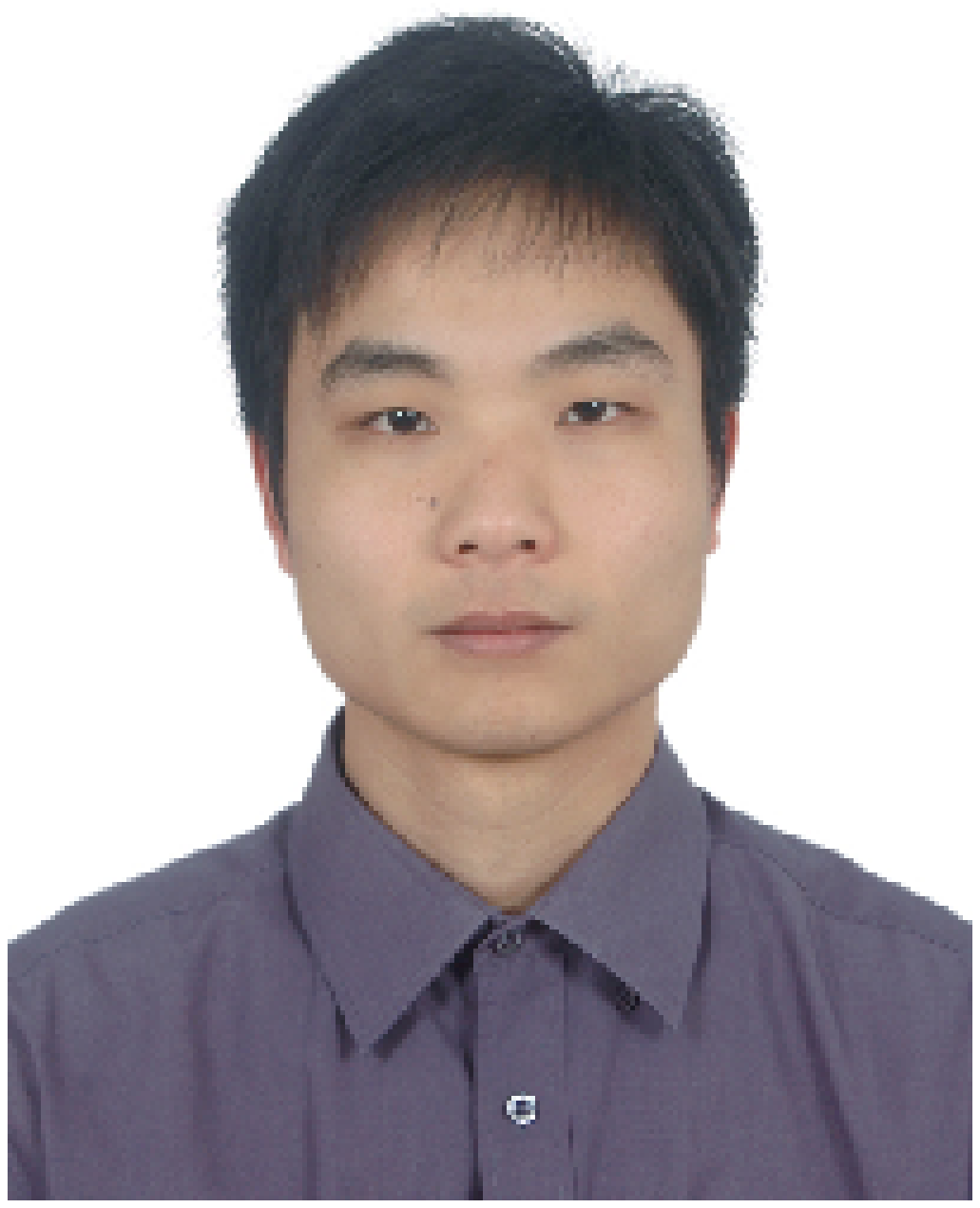}}]{Chun-Yang Zhang} received the B.S. degree in Mathematics from Beijing Normal University Zhuhai, China, in 2010 and M.S. degree in Mathematics from University of Macau, Macau, in 2012. He also received the Ph.D. degree in Computer Sciences from University of Macau, Macau, in 2015. He is currently working as an associate professor in School of Computer and Big Data at Fuzhou University. His research interests include machine learning, computer vision, computational intelligence, and big data analysis.
\end{IEEEbiography}

\begin{IEEEbiography}[{\includegraphics[width=1in,height=1.25in,clip,keepaspectratio]{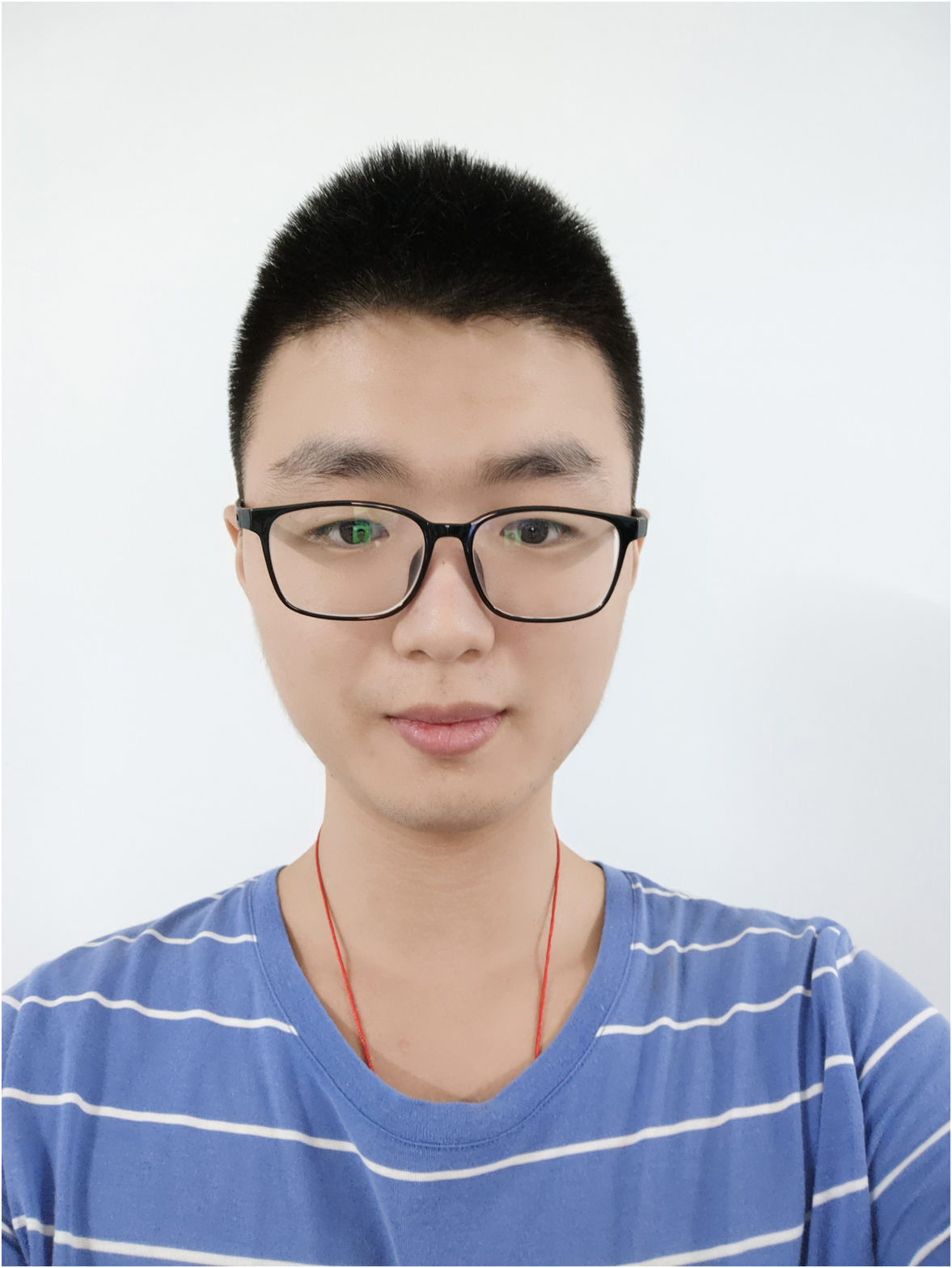}}]{Hong-Yu Yao} received the B.S. degree in computer science from Fujian Normal University, Fuzhou, China, in 2020. He is currently pursuing the graduation degree with Fuzhou University, Fuzhou, China. His research interests include unsupervised learning and graph representation learning.
\end{IEEEbiography}

\begin{IEEEbiography}[{\includegraphics[width=1in,height=1.25in,clip,keepaspectratio]{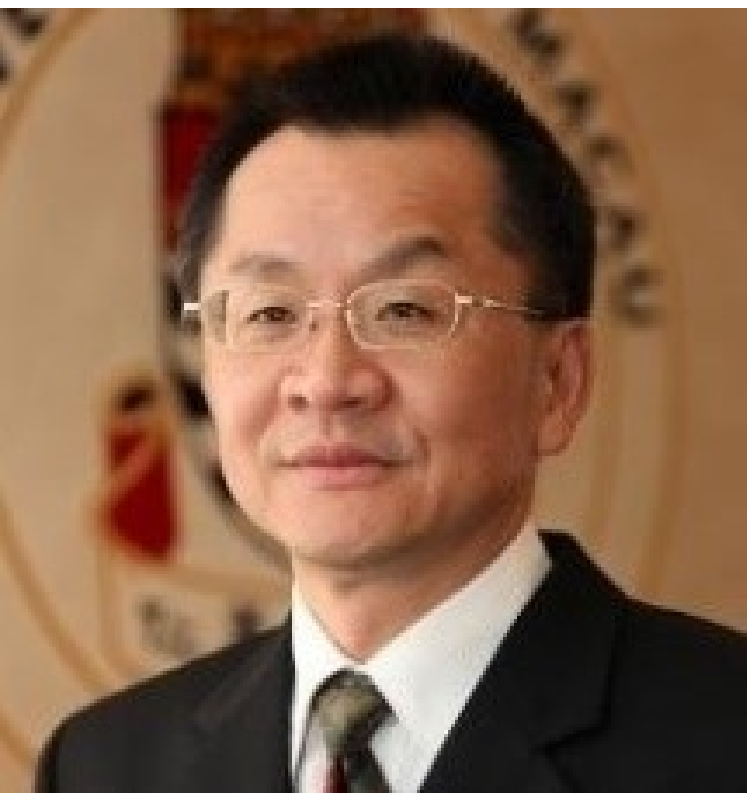}}] (S'88-M'88-SM'94-F'07) C. L. Philip Chen is the Chair Professor and Dean of the College of Computer Science and Engineering, South China University of Technology. Being a Program Evaluator of the Accreditation Board of Engineering and Technology Education (ABET) in the U.S., for computer engineering, electrical engineering, and software engineering programs, he successfully architects the University of Macau’s Engineering and Computer Science programs receiving accreditations from Washington/Seoul Accord through Hong Kong Institute of Engineers (HKIE), of which is considered as his utmost contribution in engineering/computer science education for Macau as the former Dean of the Faculty of Science and Technology. He is a Fellow of IEEE, AAAS, IAPR, CAA, and HKIE; a member of Academia Europaea (AE), European Academy of Sciences and Arts (EASA), and International Academy of Systems and Cybernetics Science (IASCYS). He received IEEE Norbert Wiener Award in 2018 for his contribution in systems and cybernetics, and machine learnings. He is also a 2018 highly cited researcher in Computer Science by Clarivate Analytics.

His current research interests include systems, cybernetics, and computational intelligence. Dr. Chen was a recipient of the 2016 Outstanding Electrical and Computer Engineers Award from his alma mater, Purdue University, after he graduated from the University of Michigan at Ann Arbor, Ann Arbor, MI, USA in 1985. He was the IEEE Systems, Man, and Cybernetics Society President from 2012 to 2013, and currently, he is the Editor-in-Chief of the IEEE Transactions on Systems, Man, and Cybernetics: Systems, and an Associate Editor of the IEEE Transactions on Fuzzy Systems, and IEEE Transactions on Cybernetics. He was the Chair of TC 9.1 Economic and Business Systems of International Federation of Automatic Control from 2015 to 2017 and currently is a Vice President of Chinese Association of Automation (CAA).
\end{IEEEbiography}

\begin{IEEEbiography}[{\includegraphics[width=1in,height=1.25in,clip,keepaspectratio]{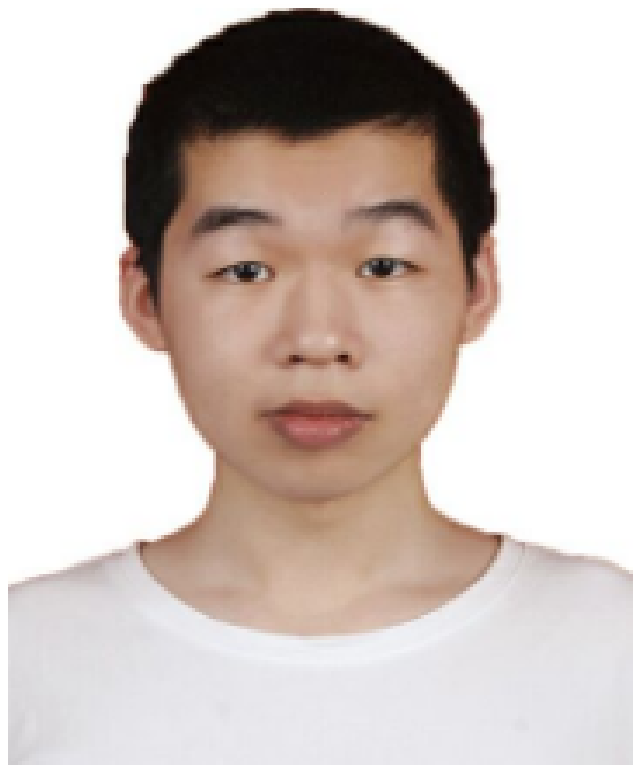}}]{Yue-Na Lin} received the B.S. degree in computer science from Fujian Normal University, Fuzhou, China, in 2020. He is currently pursuing the graduation degree with Fuzhou University, Fuzhou, China. His research interests include machine learning and big data analysis.
\end{IEEEbiography}

\end{document}